\documentclass[sigconf]{acmart}
\usepackage{bm}
\usepackage{multirow}
\usepackage{multicol}
\usepackage{stfloats}
\usepackage{caption}
\usepackage{subfigure}
\renewcommand\footnotetextcopyrightpermission[1]{}

\AtBeginDocument{%
  \providecommand\BibTeX{{%
    \normalfont B\kern-0.5em{\scshape i\kern-0.25em b}\kern-0.8em\TeX}}}

\begin{document}

\title{Auto-GNN: Neural Architecture Search of Graph Neural Networks}
\author{Kaixiong Zhou, Qingquan Song, Xiao Huang, Xia Hu}
\affiliation{Department of Computer Science and Engineering, Texas A\&M University}
\email{{zkxiong, song_3134, xhuang, xiahu}@tamu.edu}

\begin{abstract}
Graph neural networks (GNN) has been successfully applied to operate on the graph-structured data. Given a specific scenario, rich human expertise and tremendous laborious trials are usually required to identify a suitable GNN architecture. It is because the performance of a GNN architecture is significantly affected by the choice of graph convolution components, such as aggregate function and hidden dimension. Neural architecture search (NAS) has shown its potential in discovering effective deep architectures for learning tasks in image and language modeling. However, existing NAS algorithms cannot be directly applied to the GNN search problem. First, the search space of GNN is different from the ones in existing NAS work. Second, the representation learning capacity of GNN architecture changes obviously with slight architecture modifications. It affects the search efficiency of traditional search methods. Third, widely used techniques in NAS such as parameter sharing might become unstable in GNN.

To bridge the gap, we propose the automated graph neural networks (AGNN) framework, which aims to find an optimal GNN architecture within a predefined search space. A reinforcement learning based controller is designed to greedily validate architectures via small steps. AGNN has a novel parameter sharing strategy that enables homogeneous architectures to share parameters, based on a carefully-designed homogeneity definition. Experiments on real-world benchmark datasets demonstrate that the GNN architecture identified by AGNN achieves the best performance, comparing with existing handcrafted models and tradistional search methods. 
\end{abstract}



\maketitle
\footnotetext{Preprint. Under review.}
\section{Introduction}
Graph neural networks (GNN) \cite{gori2005new, scarselli2009graph} has been demonstrated that it could achieve superior performance in modeling graph-structured data, within various domains such as social media~\cite{grover2016node2vec, perozzi2014deepwalk, tang2015line, wang2016structural} and bioinformatics~\cite{zitnik2017predicting, RNN2018}.
 Following the message passing strategy~\cite{hamilton2017inductive}, GNN iteratively learns a node's embedding representations via aggregating representations of its neighbors and itself. The learned node representations could be employed by downstream machine learning algorithms to perform different tasks efficiently.
 

However, the success of GNN is accompanied with laborious work of neural architecture tuning, aiming to adapt GNN to different graph-structure data. For example, the attention heads in the graph attention networks ~\cite{velickovic2017graph} are selected carefully for citation networks and protein-protein interactions.
GraphSAGE~\cite{hamilton2017inductive} has been shown to be sensitive to hidden dimensions. 
These handcrafted architectures not only require extensive search in the design space through many trials, but also tend to obtain suboptimal performance when they are transferred to other graph-structured datasets. 
Naturally, there is a raising demand for automated GNN search to identify the optimal architecture for different real-world scenarios.

Recently, neural architecture search (NAS) has attracted increasing research interests \cite{elsken2018neural}. Its goal is to find the optimal neural architecture in the predefined search space to maximize model performance on a given task. The deep architectures discovered by NAS algorithms have outperformed the handcrafted ones at the domains including image classification \cite{zoph2016neural, zoph2018learning, liu2017hierarchical, pham2018efficient, jin2018auto, luo2018neural, liu2018progressive, liu2018darts, xie2019exploring, kandasamy2018neural}, semantic image segmentation \cite{liu2019auto}, and image generation \cite{wang2019agan}. Motivated by the success of NAS, we extend NAS studies beyond the image domains to node classification.

However, the direct application of NAS algorithms to find GNN architectures is non-trivial due to three major challenges as follows. {\textit{First, the search space of GNN architecture is different with the ones in existing NAS work}}.
Taking the search of convolutional neural network (CNN) based architectures~\cite{zoph2016neural} as an example, the convolution operation is specified only by the kernel size. In contrast, the message-passing based graph convolution in GNN is described by a sequence of actions, including aggregation, combination, and activation. 
{\textit{Second, the traditional controller is inefficient to discover the potentially well-performed GNN architecture}}. It is because the representation learning capacity of GNN architecture varies significantly with slight architecture modification. In contrast, the widely-used controller samples a complete neural architecture at each search step, and gets update after validating the new architecture. It would be hard for the traditional controller to learn the following causality: which part of the architecture modification improves or degrades the model performance. 
For example, the traditional controller changes the action sequence in new GNN architecture, and cannot distinguish the improvement brought only by replacing the aggregate function of max pooling with summation~\cite{howpowerful}.
{\textit{Third, the widely-used techniques in NAS such as parameter sharing is not suitable to GNN architecture}}. The parameter sharing transfers weight trained from one architecture to another one, aiming to avoid training from scratch. But it would lead to unstable training when sharing parameters among heterogeneous GNN architectures.
We say that two neural architectures are heterogeneous if they have different shape of trainable weight or output statistics. 
The weights of architectures with different shapes cannot be directly shared. Output statistics \cite{guo2019single} is defined as the mean, variance, or interval of the output value in each graph convolutional layer of GNN architecture. Suppose that we have parameters deeply trained in a layer with $\mathrm{Sigmoid}$ activation function, bounding the output within interval [$0, 1$]. If we transfer the parameter to another layer possessing $\mathrm{Linear}$ function, the output value may be too large to be backpropagated steadily in the gradient decent optimizer. 


To tackle the abovementioned challenges, we investigate the automated graph neural architecture search problem. Specifically, it could be separated as two research questions. (i) How to define the search space of GNN architecture, and explore it efficiently? (ii) How to constrain the parameter sharing among the heterogeneous GNN architectures to make training more stably? In summary, our major contributions are described below. 
\begin{itemize}
    \item We formally define the neural architecture search problem tailored to graph neural networks.
    
    
    \item We design a more efficient controller by considering a key property of GNN architecture--the variation of representation learning capacity with slight architecture modification.
    \item We define the heterogeneous GNN architectures in the context of parameter sharing, to train the architecture more stable with shared weight. 
    \item The experiments show that the discovered neural architecture consistently outperforms state-of-the-art handcrafted models and other search methods.
\end{itemize}

\section{Problem Statement}
We formally define the graph neural architecture search problem as follows.
Given search space $\mathcal{F}$, training set $D_{\mathrm{train}}$, validation set $D_{\mathrm{valid}}$ and evaluation metric $M$, we aims to find the optimal GNN architecture $f^* \in \mathcal{F}$ accompanied with the best metric $M^*$ on set $D_{\mathrm{valid}}$. Mathematically, it is written as follows.
\begin{equation}
    \label{equ:NAS}
    \begin{split}
        f^* & = \mathrm{argmax}_{f\in \mathcal{F}}\ M(f(\theta^*), D_{\mathrm{valid}}) \\
        \theta^* & = \mathrm{argmin}_{\theta}\ L(f(\theta), D_{\mathrm{train}}).
    \end{split}
\end{equation}
$\theta^*$ denotes the parameter learned for architecture $f$ and $L$ denotes the loss function. Metric $M$ could be represented by F1 score or accuracy for node classification task. The characteristics of GNN search problem could be viewed from three aspects. First, search space $\mathcal{F}$ is constructed based graph convolutions. Second, an efficient controller is required to consider the relationship between model performance and slight architecture modification in GNN. Third, the parameter sharing needs to promise weight could be transferred stably among heterogeneous GNN architectures. 

\begin{figure*}
    \centering
    \includegraphics[width=0.95\textwidth]{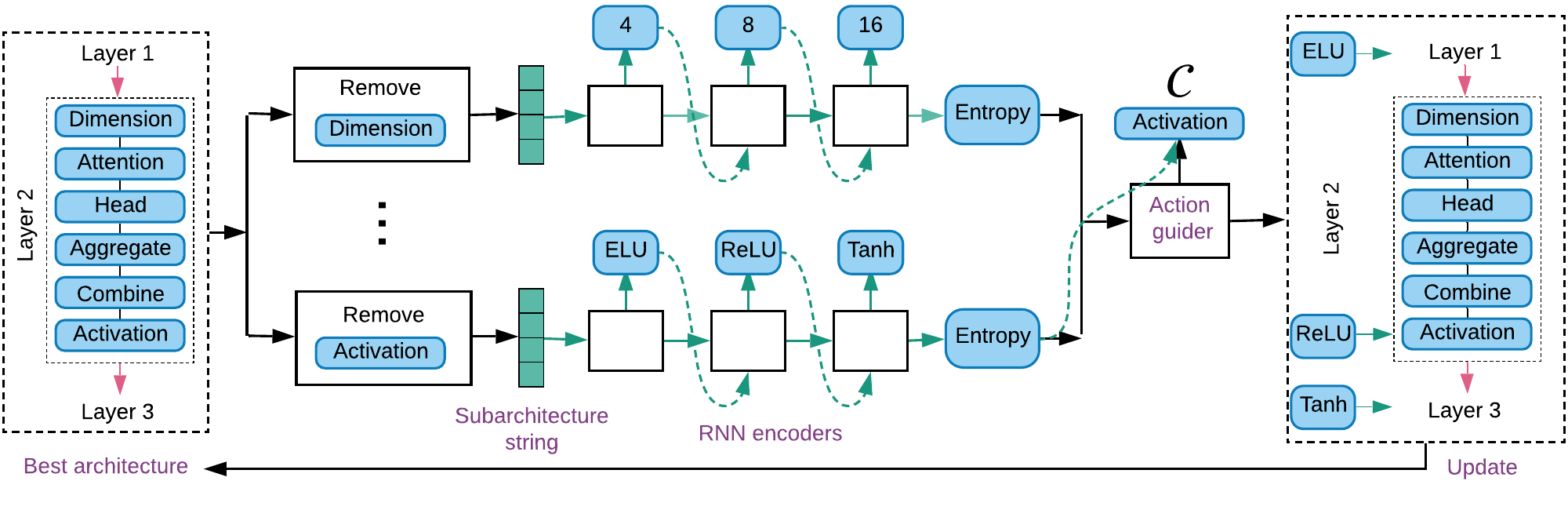}\\
    \vspace{-10pt}
    \caption{Illustration of AGNN with $3$-layer GNN search. Controller takes the best architecture found so far as input, and removes one of the six classes in turns to generate six subarchitectures. Their strings are fed to RNN encoders to determinate the best alternative action for the missing class. We select the new best architecture from all completed subarchitectures, based the accompanied decision entropy.
    Herein action guider selects class list $\mathcal{C} = \{\mathrm{Activation\ function}\}$. The retained architecture is modified via replacing activation functions with $\mathrm{ELU}$, $\mathrm{ReLU}$, and $\mathrm{Tanh}$, in all $3$ graph convolutional layers, respectively.}
    \label{fig:AGNN}
\end{figure*}

We propose an efficient and effective framework named AGNN to handle the GNN search problem. Figure~\ref{fig:AGNN} illustrates its core idea via a $3$-layer GNN architecture search example.
In the search space, each graph convolutional layer is specified by an action sequence as listed in the left box. There are totally six action classes, which cover a wide-variety of state-of-the-art GNN models.
Instead of resampling a completely new neural architecture, we have independent RNN encoders to decide the new action for each class, e.g., the hidden dimension and activation function. Controller keeps the best architecture found so far, and makes slight architecture modification to it on specific classes. As shown in the right hand of figure, we change the activation functions in all $3$ layers of the retained architecture to $\mathrm{ELU}$, $\mathrm{ReLU}$ and $\mathrm{Tanh}$, respectively. In this way, we are able update each RNN encoder independently to learn the affect of specific action class to model performance. 
A tailored parameter sharing strategy is designed. It defines homogeneous GNN architectures via three constraints. Weight only shares from the homogeneous ancestor architecture, helping the offspring architecture train stably. We will update the best architecture if the offspring architecture outperforms it; otherwise, we continue the search by reusing the old best architecture. Next, we introduce the search space, controller, and parameter sharing in detail.




\section{Search Space}
In this section, we describe the designed search space for the general GNN architecture, which is composed of layers of message-passing based graph convolutions. Formally, the $k$-th layer 
\begin{equation}
    \label{eq:GNN}
    \begin{split}
        h^{(k)}_i & = \mathrm{AGGREGATE}(\{a^{(k)}_{ij} W^{(k)} x^{(k-1)}_j: j\in \mathcal{N}(i)\}), \\
        x^{(k)}_i & = \mathrm{ACT}(\mathrm{COMBINE}(W^{(k)}x^{(k-1)}_i, h^{(k)}_i)).
    \end{split}
\end{equation}
$x^{(k)}_i$ denotes the embedding of node $i$ at the $k$-th layer. $\mathcal{N}(i)$ denotes the set of nodes adjacent to node $i$. $W^{(k)}$ denotes the trainable matrix used to transform embedding dimension. $a^{(k)}_{ij}$ denotes the attention coefficient between nodes $i$ and $j$ obtained from the additional attention layer. Function $\mathrm{AGGREGATE}$ is applied to aggregate neighbor representations and prepare intermediate embedding $h^{(k)}_i$. In addition, function $\mathrm{COMBINE}$ is used to combine information from node itself as well as intermediate embedding $h^{(k)}_i$, and function $\mathrm{ACT}$ is used to activate the node embedding. Based on the message-passing graph convolutions defined in Equation (\ref{eq:GNN}), we decompose the search space into the following $6$ classes of actions:
\begin{itemize}
    \item \textbf{Hidden dimension:} Trainable matrix $W^{(k)}$ extracts representative features from embedding $x^{(k-1)}_i$ of the last layer, and maps the embedding to a $d$-dimensional space. The choice of dimension $d$ is crucial to the final node classification performance. We collect the set of dimensions that are widely adopted by existing work as the candidates, i.e., $\{$4$, $8$, $16$, $32$, $64$, $128$, $256$\}$.
    \item \textbf{Attention function:} The real-world graph-structured data could be both complex and noisy \cite{lee2018attention}, which may lead to the inefficient information aggregation. The attention mechanism helps to focus on the most relevant neighbors to improve the representative learning of node embedding. Following NAS framework in \cite{gao2019graphnas}, we collect the set of attention functions as shown in Table \ref{Tab: attention} to compute coefficient $a^{(k)}_{ij}$.
    \item \textbf{Attention head: } It is found that the multi-head attention could be beneficial to stabilize the learning process \cite{velickovic2017graph,vaswani2017attention}. We select the number of attention heads within the set: $\{$1$, $2$, $4$, $6$, $8$, $16$\}$.
    \item \textbf{Aggregate function:} As shown in \cite{howpowerful}, aggregate function is crucial to capture neighborhood structure for learning node representation. Herein GNN architecture is developed based on package Pytorch Geometric \cite{Fey/Lenssen/2019}.
    The package provides the following available aggregate functions: $\{\mathrm{SUMMATION}, \mathrm{MEAN}, \mathrm{MAXPOOLING}\}$.
    \item \textbf{Combine function:} Embeddings $W^{(k)}x^{(k-1)}_i$ and $h^{(k)}_i$ are usually concatenated to combine information from node itself and neighbors. A differentiable function could then be applied to enhance the node representation learning. We design to select from two types of combine functions: $\{\mathrm{IDENTITY}, \mathrm{MLP}\}$. Herein MLP is a $2$-layer perceptron with a fixed hidden dimension of $128$. 
    \item \textbf{Activation function: } The set of available activation functions in our AGNN is listed as follows: $\{\mathrm{Sigmoid}, \mathrm{Tanh}, \mathrm{ReLU}, \\ \mathrm{Linear}, \mathrm{Softplus}, \mathrm{LeakyReLU}, \mathrm{ReLU6}, \mathrm{ELU\}}$
\end{itemize}


Note that a wide-variety of state-of-the-art model fall into the above message-passing based GNN architecture, including Chebyshev \cite{defferrard2016convolutional}, GCN \cite{kipf2016semi}, GraphSAGE \cite{hamilton2017inductive}, GAT \cite{velickovic2017graph} and LGCN \cite{gao2018large}. We apply the fixed skip connection as those in \cite{kipf2016semi, velickovic2017graph}. The skip connection action could be easily incorporated into search space if necessary. Equipped with the above design, a GNN architecture could be specified by a string of length $6n$, where $n$ denotes the number of graph convolutional layers. For each layer, cardinalities of the above six action classes are $7$, $7$, $6$, $3$, $2$, $8$, respectively, which provides $7\times 7\times 6\times 3\times 2\times 8 = 14112$ possible combinations in total. Suppose we target at searching a three-layer GNN architecture, i.e., $n=3$, which is commonly accepted in GNN models. The number of unique 
architectures within our search space is $(14112)^3 \approx 2.8\times 10^{12}$, which is quite large and multifarious.




\begin{table}
\centering
\caption{The set of attention functions, where symbol $||$ denotes the concatenation operation, $\vec{a}$, $\vec{a}_l$ and $\vec{a}_r$ denote the trainable vectors, and $W_{G}$ denotes the trainable matrix. }
\label{Tab: attention}
\begin{tabular}{c|c}
\toprule
   Attention Mechanisms &  Equations \\
   \hline
   \hline
   $\mathrm{CONSTANT}$ & $1$ \\ \hline
   $\mathrm{GCN}$ & $\frac{1}{\sqrt{|\mathcal{N}(i)||\mathcal{N}(j)|}}$ \\ \hline
   $\mathrm{GAT}$ & $\mathrm{LeakyReLU}(\vec{a}(W^{(k)}x^{(k-1)}_i||W^{(k)}x^{(k-1)}_j))$ \\ \hline
   $\mathrm{SYM}$-$\mathrm{GAT}$ & $a^{(k)}_{ij}+a^{(k)}_{ji}$ $\mathrm{based\ on\ GAT}$ \\  \hline
   $\mathrm{COS}$ & $\vec{a}(W^{(k)}x^{(k-1)}_i||W^{(k)}x^{(k-1)}_j)$ \\ \hline
   $\mathrm{LINEAR}$ & $\mathrm{tanh}(\vec{a_l}W^{(k)}x^{(k-1)}_i+\vec{a_r}W^{(k)}x^{(k-1)}_i)$ \\ \hline
   $\mathrm{GERE}$-$\mathrm{LINEAR}$ & $W_{G}\mathrm{tanh}(W^{(k)}x^{(k-1)}_i+W^{(k)}x^{(k-1)}_i)$ \\
\bottomrule
\end{tabular}
\end{table}

\section{Reinforced Conservative Controller}


In this section, we elaborate the proposed controller aiming to search GNN architecture efficiently. The controller framework is built up upon RL-based exploration guided with conservative exploitation. In traditional RL-based NAS, RNN is applied to specify the variable-length neural architecture, and generate a new candidate architecture at each search step. All of the action components in the neural architecture will be resampled and replaced with the new ones. After validating the new architecture, a scalar reward is made use to update the RNN. However, it could be problematic to directly apply this traditional controller to find potentially well-performed GNN architectures. The main reason is that the representation learning capacity of GNN architecture varies significantly with slight modification of some action classes. Taking the aggregate function as example, the classification performance of GNN architecture may improve by only replacing the function of max pooling with summation~\cite{howpowerful}. It would be hard for the conventional controller to learn about which part of architecture modification contributes more to the performance improvement.





In order to tackle the above challenge, we propose a new searching algorithm named reinforced conservative neural architecture search (RCNAS). It consists of three components: (1) A conservative explorer, which screens out the best architecture found so far. (2) A guided architecture modifier, which slightly mutates certain actions in the retained best architecture. (3) A reinforcement learning trainer that learns the architecture modification causality. In the following, we introduce the details of these three components.

\subsection{Conservative Explorer}
As the key exploitation component, the conservative explorer is applied to maintain the best neural architecture found so far. In this way, the following architecture modification is performed based on a reliable well-performed architecture, which ensures a fast exploitation towards better architectures among the offsprings generated from slight architecture modification. If the offspring architecture outperforms its parent one, we will update the best neural architecture; otherwise, the best one will be kept and reused to generate the next offspring architecture. In practice, multiple starting points could be randomly initialized to enhance the exploration ability and avoid trapping in local minimums.


\subsection{Guided Architecture Modifier}


The main role of the guided architecture modifier is to modify the best architecture found so far via selecting and mutating the action classes that wait for exploration. As shown in the right hand of Figure \ref{fig:AGNN}, assume the class of activation function is selected. Correspondingly, the actions of activation function in the $3$-layer GNN architecture are resampled and changed to $\mathrm{ELU}$, $\mathrm{ReLU}$ and $\mathrm{Tanh}$, respectively. 
This will facilitate controller to learn the affect of architecture modification on specific action class. 

To be specific, the architecture modification is realized by three steps: (1) For each class, an independent RNN encoder decides a sequence of new actions. (2) An action guider receives the decision entropy and selects the action classes to be modified. (3) An architecture modification generates the final offspring architecture. Details are introduced as follows.


 

\subsubsection{\textbf{RNN Encoders}:} As shown in Figure \ref{fig:AGNN}, for each class, an independent RNN encoder is implemented to decide a sequence of new actions. First, a subarchitecture string of length $5n$ is generated by removing $n$ actions of concerned class. For example, considering the $3$-layer neural architecture in Figure \ref{fig:AGNN}, the subarchitecture of class activation function is obtained by removing activations existing in all $3$ convolutional layers of the best architecture. Second, following an embedding layer, the subarchitecture string is taken as input to RNN encoder. This string represents the input status that asks for action padding of concerned class. Third, RNN encoder iteratively outputs the candidate action; and the output is then fed into next step as input. Note that the candidate action is sampled by feeding hidden state $h_i$ into a softmax classifier. The length of each RNN encoder is $n$, coupling with the number of layers to be searched in the architectures.




\subsubsection{\textbf{Action Guider}:} It is responsible to receive the decision entropy of each RNN encoder, and select some classes to be modified on the retained architecture. Consider the decision entropy of class $c$. At step $i$ of RNN encoder, hidden state $h_i$ is fed into the softmax classifier, and a probability vector $\bm{\vec P}_i$ is given as output. 
The $j$-th element $P_{ij}$ represents the probability of sampling action $j$. The decision entropy of class $c$ is then given by: $E_c \triangleq \sum_{i=1}^{n}\sum_{j=1}^{m_c} -P_{ij}\log P_{ij}$, where $m_c$ denote the action cardinality of class $c$. Decision entropy $E_c$ represents the uncertainty of current subarchitecture to explore along action class $c$.

Given decision entropy list $\{E_1, \cdots, E_6\}$ of the six action classes, the action guider samples classes $\mathcal{C} = \{c_1, \cdots, c_s\}$ with size $s$, which would be used to modify network architecture. For example, class activation function is selected as shown in Figure \ref{fig:AGNN}, where $\mathcal{C} = \{\mathrm{Activation\ function}\}$, $s=1$. The larger the decision entropy $E_c$ is, the larger the probability class $c$ are desired to be sampled. The action guider help controller search the potential networks along the direction with most uncertainty, which performs similar to the Bayesian optimization method \cite{jin2018auto}. 

\subsubsection{\textbf{Architecture Modification}:} The best architecture found so far is modified via replacing the corresponding actions of each class in list $\mathcal{C}$. In Figure \ref{fig:AGNN}, action list $\{\mathrm{ELU}, \mathrm{ReLU}, \mathrm{Tanh}\}$ is applied to replace the activation functions existing in all of the $3$ graph convolutional layers. When list $\mathcal{C}$ includes only one class, we modify the retained neural architecture at a minimum level. If size $s=6$, our controller resamples actions in the whole architecture similar to the traditional controller.

\subsection{Reinforcement Learning Trainer}
We use the REINFORCE rule of policy gradient \cite{sutton2000policy} to update parameters $\theta_c$ for RNN encoder of class $c \in \mathcal{C}$. Let $\{a_1, \cdots, a_n\}$ denote the decided action list of class $c$. 
We have the following update rule \cite{zoph2016neural}:
\begin{equation}
    \label{eq:reinforce}
    \nabla_{\theta_c} J(\theta_c) = \sum_{t=1}^{n}\mathbb{E}[(R_c - b_c) \nabla_{\theta_c}\log\mathbb{P}(a_t|a_{t-1}; \theta_c)],
\end{equation}
where $R_c$ denotes the reward for taking decisions $\{a_1, \cdots, a_n\}$ of class $c$, and $b_c$ denotes the baseline of class $c$ for variance reduction. Let $M_b$ and $M_o$ denote the model performances of the best architecture found so far and its offspring one, respectively. We propose the following reward shaping: $R_c \triangleq M_o - M_b$, which represents the performance variation brought by modifying the retained architecture on the class $c$.

\section{Constrained Parameter Sharing}

Compared to training from scratch, parameter sharing reduces the computation cost via forcing the offspring architecture to share weight already trained well in the ancestor architecture. However, the traditional strategy cannot be directly applied to share weight among the heterogeneous GNN architectures. We say that two neural architectures are heterogeneous if they have different shapes of trainable weight or output statistics. 
First, the distinct weight shape in the offspring architecture prevents the direct transfer from an ancestor architecture.
Second, weight is deeply trained and coupled in the ancestor architecture. The shared weight from heterogeneous architecture with different output statistics may lead to output explosion and unstable training \cite{guo2019single}. Consider the output intervals of activation functions $\mathrm{Sigmoid}$ and $\mathrm{Linear}$, which are given by [$0, 1$] and [$-\infty, +\infty$], respectively. The shared wight is unsuitable to the architecture possessing function $\mathrm{Linear}$ when it is transferred from the one possessing function $\mathrm{Sigmoid}$. Third, the shared weights in the connection layer may not be effective and adaptive to the offspring architecture immediately. The connection layer is given by the batch normalization or skip connection, and may be uncoupled to the offspring architecture. 

\begin{figure}
    \centering
    \includegraphics[width=0.5\textwidth]{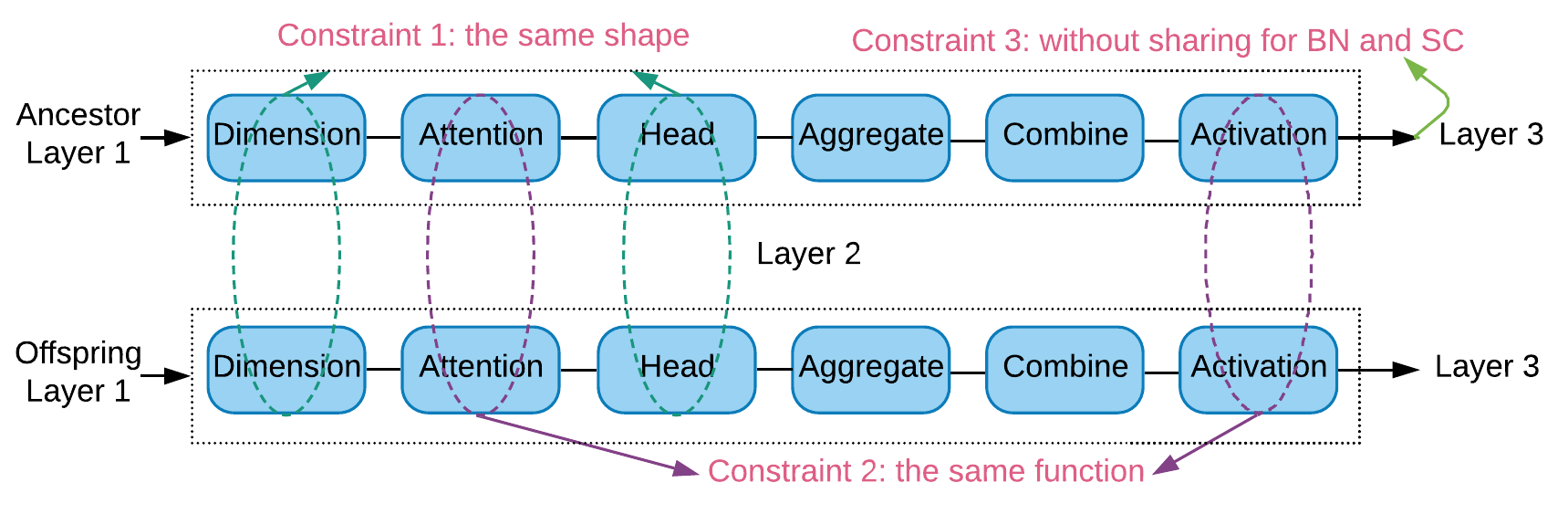}\\
    \vspace{-10pt}
    \caption{An illustration of the constrained parameter sharing strategy between the ancestor and offspring architectures. The trainable parameter of a convolutional layer could only be shared when they have the same weight shape (constraint $1$), attention and activation functions (constraint $2$). Constraint $3$ removes the parameter sharing for batch normalization (BN) and skip connection (SC).}
    \label{fig:param-share}
\end{figure}

To tackle the above challenges, we propose the constrained parameter sharing strategy to limit how the offspring architecture inheriting parameter from ancestor architectures found before. 
As shown in Figure \ref{fig:param-share}, we explain the three constraints as follows:
\begin{itemize}
    \item The ancestor and offspring architectures have the same shape of input and output tensors for the graph convolutional layer. Based on the graph convolutions defined in Equation (\ref{eq:GNN}), both trainable matrix $W^{(k)}$ and transform weight used in the attention function could be shared directly only if they have the same shape. 
    \item The ancestor and offspring architectures have the same attention function and activation function for the graph convolutional layer. The attention function defines the neighbor information to be aggregated, and the activation function squashes the output to a specific interval. Hence both attention function and activation function greatly determines the output statistics of a graph convolutional layer. It is expected to void output explosion and improve the training stability via sharing parameter from homogeneous architecture with similar output statistics. 
    \item The parameters of batch normalization (BN) and skip connection (SC) will not be shared. It is because we do not know the exact output statistics of each layer in the offspring architecture in advance. The shared parameters of BN and SC may cannot bridge the two successive layers well. We train the whole offspring architecture with a few epochs (e.g., $5$ or $20$ epochs in our experiment), to adapt these parameters to the new architecture. 
\end{itemize}

\section{Experiments}
We apply our method to find the optimal GNN architecture given the node classification task, to answer the following four questions:
\begin{itemize}
    \item \textbf{Q1:} How does the GNN architecture discovered by AGNN compare with state-of-the-art handcrafted architectures and the ones searched by other methods?
    \item \textbf{Q2:} How does the search efficiency of RCNAS controller compare with those of other search methods?
    \item \textbf{Q3:} Whether or not the constrained strategy shares weight effectively, to help the offspring architecture achieve good classification performance?
    \item \textbf{Q4:} How does different scales of architecture modification affect the search efficiency of the RCNAS controller?
\end{itemize}
More details about the datasets, baseline methods, experimental configuration and results are introduced as follows.

\begin{table}[!htbp]
\setlength{\tabcolsep}{4pt}
  \caption{Statistics of  datasets Cora, Citeseer, Pubmed, and PPI \cite{velickovic2017graph, gao2018large}, where T and I denote the transductive and inductive learning, respectively.}
  \small
  \label{tab:dataset}
  \begin{tabular}{c|cccc}
    \toprule
     & Cora & Citeseer & Pubmed & PPI\\
    \midrule
    Setting & T & T & T & I \\
    \#Nodes & $2708$ & $3327$ & $19717$ & $56944$ \\
    \#Features & $1433$ & $3703$ & $500$ & $50$ \\
    \#Classes& $7$ & $6$ & $3$ & $121$ \\
    \#Training Nodes & $140$ & $120$ & $60$ & $44906$ ($20$ graphs) \\
    \#Validation Nodes & $500$ & $500$ & $500$ &  $6514$ ($2$ graphs) \\
    \#Testing Nodes & $1000$ & $1000$ & $1000$ & $5524$ ($2$ graphs) \\
  \bottomrule
\end{tabular}
\end{table}

\subsection{Datasets} 
We consider both transductive and inductive learning settings for the node classification task. Under the transductive learning, the unlabeled data used for validation and testing are accessible during training. This means the training process could make use of the complete graph structure and node features, except for node labels on the held-out validation and testing sets. Under the inductive learning, the training process has no idea about the graph structure and node features on both validation and testing sets.

We utilize Cora, Citeseer and Pubmed \cite{sen2008collective} for the transductive learning, and use PPI for the inductive learning \cite{zitnik2017predicting}. These benchmark datasets are commonly used for studying the node classification task. The dataset statistics is given in Table \ref{tab:dataset}. The three datasets evaluated under transductive learning are citation networks, where node corresponds to document and edge corresponds to citation relation. Node feature is given by bag-of-words representation of a document, and each node is associated with a class label. Following the same experimental setting as those in baseline methods, we allow for $20$ nodes per class to be used for training, and use $500$ and $1000$ nodes for validation and testing, respectively. PPI dataset evaluated under inductive learning consists of graphs corresponding to different human tissues. There are $50$ features for each node, including the positional gene sets, motif gene sets and immunological signatures. Each node has several labels simultaneously collected from total of $121$ classes. We use $20$ graphs for training, $2$ graphs for validation and $2$ graphs for testing. The model metric is given by classification accuracy and micro-averaged F1 score for transductive learning and inductive learning, respectively.

\subsection{Baseline Methods}
In order to evaluate our method designed specifically for finding GNN architecture, we consider the baselines of both state-of-the-art handcrafted architectures as well as other NAS approaches. 
\begin{itemize}
    \item \textbf{Handcrafted architectures:} Herein we only consider the message-passing based GNNs as shown in Equation (\ref{eq:GNN}) for fair comparison, except the one combined with pooling layer. The following baseline methods are included: Chebyshev \cite{defferrard2016convolutional}, GCN \cite{kipf2016semi}, GraphSAGE \cite{hamilton2017inductive}, GAT \cite{velickovic2017graph}, LGCN \cite{gao2018large}. Note that both Chebyshev and GCN perform information aggregation based on the Laplacian or adjacent matrix of the complete graph. Hence they are only evaluated under the transductive learning setting. Baseline GraphSAGE aggregates information via sampling neighbors pf fixed size, which will be compared only under the inductive learning setting. We consider a variety of GraphSAGE possessing different aggregate functions, including GraphSAGE-GCN, GraphSAGE-mean, GraphSAGE-pool and GraphSAGE-LSTM.
    \item \textbf{NAS approaches:} We compare with the previous NAS approaches based on reinforcement learning and random search. The former one utilizes RNN to sample the whole neural architecture, and applies reinforcement rule to update controller. GraphNAS proposed in \cite{gao2019graphnas} applies this approach directly to search GNN architecture. The later one samples architecture randomly, serving as baseline to evaluate the efficiency of our controller.    
\end{itemize}

\subsection{Training Details}
We train the sampled neural architecture on the training set, and update the controller via receiving reward from the validation set. Following the model configurations in baselines \cite{velickovic2017graph, gao2018large}, the training experiments are set up according to transductive learning and inductive learning, respectively. We have an unified model configuration of controller. More details about our experimental procedure are introduced as follows. 

\subsubsection{\textbf{Transductive Learning}}
Herein we explore a two-layer GNN architecture in the predefined search space. Except that the neural architecture is updated iteratively during the search progress, we have the same training environment to those in the baselines. To deal with the issue of small training set, we apply L2 regularization with $\lambda = 0.0005$. Dropout rate of $0.6$ is applied to both layers’ inputs as well as the attention coefficients during training. For Pubmed dataset, L2 regularization is strengthened to $\lambda = 0.001$. 

Foe each sampled architecture, weight is initialized using Glorot initialization \cite{glorot2010understanding} and trained with Adam optimizer \cite{kingma2014adam} to minimize the cross-entropy loss. We set the initial learning rate of $0.01$ for Pubmed and $0.005$ for Cora and Citeseer. We have two different settings to train a new offspring architecture: with parameter sharing and without weight sharing. The former one has a small warm-up epochs of $20$, while the later one has $200$ training epochs.

\begin{table*}[t!]
  \centering
  \caption{Test performance comparison for architectures under the transductive learning setting: the state-of-the-art handcrafted architectures, the optimal ones found by NAS baselines, the optimal ones found by AGNN.}
  \label{tab: test-compare-transdutive}
  \begin{tabular}{l|c|c|cc|cc|cc}
    \toprule
    \multirow{2}*{\textbf{Baseline Class}} & \multirow{2}*{\textbf{Model}} & \multirow{2}*{\textbf{\#Layers}} & \multicolumn{2}{c}{\textbf{Cora}} & \multicolumn{2}{c}{\textbf{Citeseer}} & \multicolumn{2}{c}{\textbf{Pubmed}} \\
    \cline{4-9}
    & & & \#Params & Accuracy & \#Params & Accuracy & \#Params & Accuracy \\
    \hline
    \hline
     & Chebyshev & $2$ & $0.09$M & $81.2$\% & $0.09$M & $69.8$\% & $0.09$M & $74.4$\% \\
    \textbf{Handcrafted} & GCN & $2$ & $0.02$M &  $81.5$\% & $0.05$M &  $70.3$\%  & $0.02$M &  $79.0.5$\% \\
    \textbf{Architectures} & GAT & $2$ & $0.09$M & $83.0\pm 0.7\%$ & $0.23$M & $72.5\pm 0.7\%$  & $0.03$M & $79.0\pm 0.3\%$\\
    & LGCN & $3\sim 4$ & $0.06$M & $83.3\pm 0.5\%$ & $0.05$M & $73.0\pm 0.6\%$ & $0.05$M & $79.5\pm 0.2\%$ \\      
    \hline
    \multirow{4}*{\textbf{NAS Baselines}} & GraphNAS-w/o share & $2$ & $0.09$M & $82.7\pm 0.4\%$ & $0.23$M & $73.5\pm 1.0\%$ & $0.03$M & $78.8\pm 0.5\%$ \\
    & GraphNAS-with share & $2$ & $0.07$M & $83.3\pm 0.6\%$ & $1.91$M & $72.4\pm 1.3\%$ & $0.07$M & $78.1\pm 0.8\%$ \\
    & Random-w/o share & $2$ & $0.37$M & $81.4\pm 1.1\%$ & $0.95$M & $72.9\pm 0.2\%$ & $0.13$M & $77.9\pm 0.5\%$ \\
    & Random-with share & $2$ & $2.95$M & $82.3\pm 0.5\%$ & $0.95$M & $69.9\pm 1.7\%$ & $0.13$M & $77.9\pm 0.4\%$ \\
    \hline
    \multirow{2}*{\textbf{AGNN}} & AGNN-w/o share & $2$ & $0.05$M & $\bm{83.6\pm 0.3\%}$ & $0.71$M & $\bm{73.8\pm 0.7\%}$ & $0.07$M & $\bm{79.7\pm 0.4\%}$ \\
     & AGNN-with share & $2$ & $0.37$M & $82.7\pm 0.6\%$ & $1.90$M & $72.7\pm 0.4\%$ & $0.03$M & $79.0\pm 0.5\%$ \\
  \bottomrule
  \end{tabular}
\end{table*}

\subsubsection{\textbf{Inductive Learning}}
Herein we explore a three-layer GNN architecture. The skip connection between the intermediate graph convolutional layers is included to improve the representation learning. Since dataset PPI is sufficiently large for training, the L2 regularization and random dropout are removed from GNN model. The batch size of $2$ graphs is employed during training. 

We have the same parameter initialization and optimizer as the transductive learning. The initial learning rate is set to $0.005$. The warm-up epoch number is $5$ under the setting with parameter sharing, and it is $20$ under the setting without parameter sharing.

\subsubsection{\textbf{Controller}}
For each action class, RNN encoder is realized by an one-layer LSTM with $100$ hidden units. Weights are initialized uniformly in [$-0.1, 0.1$], and trained with Adam optimizer at a learning rate of $3.5\times 10^{-4}$. Following the controller configurations in the previous NAS work, we use a $\mathrm{tanh}$ constant of $2.5$ and a sample temperature of $5.0$ to the hidden output. Totally $1000$ architectures are explored iteratively during the search progress, and evaluated to obtain reward for updating controller. Reward to the policy gradient is given by the following combination: the validation performance and the controller entropy weighted by $1.0\times 10^{-4}$. 

\begin{table}
\setlength{\tabcolsep}{0.8pt}
  \centering
  \caption{Test performance comparison of our AGNN to state-of-the-art handcrafted architectures and other search approaches under the inductive learning setting.}
  \label{tab: test-compare-inductive}
  \begin{tabular}{l|c|c|cc}
    \toprule
    \textbf{Baseline} & \multirow{2}*{\textbf{Model}} & \multirow{2}*{\textbf{Layers}} & \multicolumn{2}{c}{\textbf{PPI}} \\
    \cline{4-5}
    \textbf{Class} & & & Params & F1 score \\
    \hline
    \hline
     & GraphSAGE-GCN & $2$ & $0.11$M & $0.500$ \\
     & GraphSAGE-mean & $2$ & $0.11$M & $0.598$ \\
    \textbf{Hand-} & GraphSAGE-pool & $2$ & $0.36$M & $0.600$ \\
    \textbf{crafted} & GraphSAGE-LSTM & $2$ & $0.39$M & $0.612$ \\
    & GAT & $3$ & $0.89$M & $0.973\pm 0.002$ \\
    & LGCN & $4$ & $0.85$M & $0.772\pm 0.002$ \\
    \hline
     & GraphNAS-w/o share & $3$ & $4.1$M & $0.985\pm 0.004$ \\
    \textbf{NAS} & GraphNAS-with share & $3$ & $1.4$M & $0.960\pm 0.036$ \\
    \textbf{Baselines} & Random-w/o share & $3$ & $1.4$M & $0.984\pm 0.004$ \\
    & Random-with share & $3$ & $1.4$M & $0.977\pm 0.011$ \\
    \hline
    \multirow{2}*{\textbf{AGNN}} & AGNN-w/o share & $3$ & $4.6$M & $\bm{0.992\pm 0.001}$ \\
     & AGNN-with share & $3$ & $1.6$M & $0.991\pm 0.001$ \\
  \bottomrule
  \end{tabular}
\end{table}
\subsection{Results}
In this section, we show the comparative evaluation experiments to answer the above four research questions. 
\subsubsection{\textbf{Test Performance Comparison}}
We compare the architecture discovered by our AGNN with the handcrafted ones and those found by other search methods, aiming to provide positive answer for research question $\textbf{Q1}$. Considering the architecture modification in AGNN, the default size $s$ of class list $\mathcal{C}$ is set to $1$. All of NAS approaches find the optimal architecture achieving the best performance on the separate held-out validation set. Then, it is evaluated on the testing set only once. Two comprehensive lists of architecture information and model performance are presented in Tables \ref{tab: test-compare-transdutive} and \ref{tab: test-compare-inductive} for transductive learning and inductive learning, respectively. The test performance of NAS approaches is averaged via randomly initializing the optimal architecture $5$ times, and those of handcrafted architectures are reported directly from their papers. 

As can be seen from Tables \ref{tab: test-compare-transdutive} and \ref{tab: test-compare-inductive}, the neural architecture discovered by AGNN outperforms the handcrafted ones and other search methods. Compared with the handcrafted architectures, the discovered models generally improve the classification performance accompanied with the increment of parameter size. During the search process, the larger ones of attention head and hidden dimension are explored to improve the representation learning capacity of GNN. The whole neural architecture is sampled and reconstructed in GraphNAS and random search at each step, similar to the previous NAS frameworks. In contrast, our AGNN explores the offspring architecture via only modifying specific action class. The best architecture are retained to provide a good start for architecture modification. This will facilitate the controller to learn the causality between architecture modification and model performance variation, and find the better architecture more potentially. 

It is observed that the architectures found without parameter sharing generally outperform the ones found with parameter sharing. It is because the shared parameter may be  uncoupled to the offspring architecture, although several epochs are applied to warm up. Running on a single Nvidia GTX 1080Ti GPU, it takes about $0.5$ GPU days to find the best architecture without parameter sharing, which is a few times that with parameter sharing. There is a trade-off between model performance and computation time cost. 

\begin{figure*}[t]
\centering
\hspace{-.3cm}
\subfigure[PPI]{
\begin{minipage}{4.27cm}
\centering
\includegraphics[width=4.8cm]{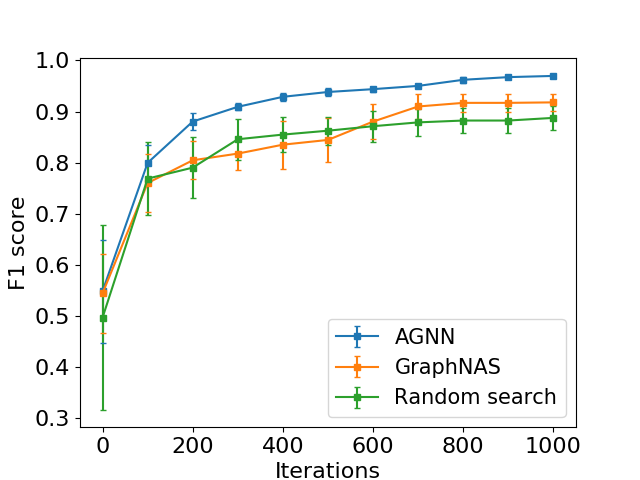}
\end{minipage}}
\subfigure[Cora]{
\begin{minipage}{4.27cm}
\centering
\includegraphics[width=4.8cm]{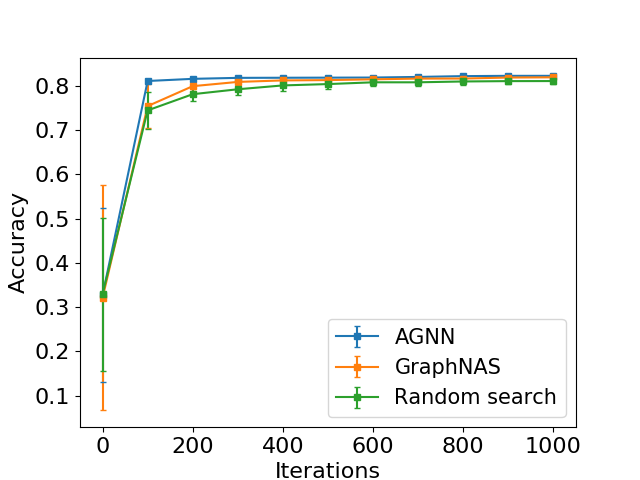}
\end{minipage}}
\subfigure[Citeseer]{
\begin{minipage}{4.27cm}
\centering
\includegraphics[width=4.8cm]{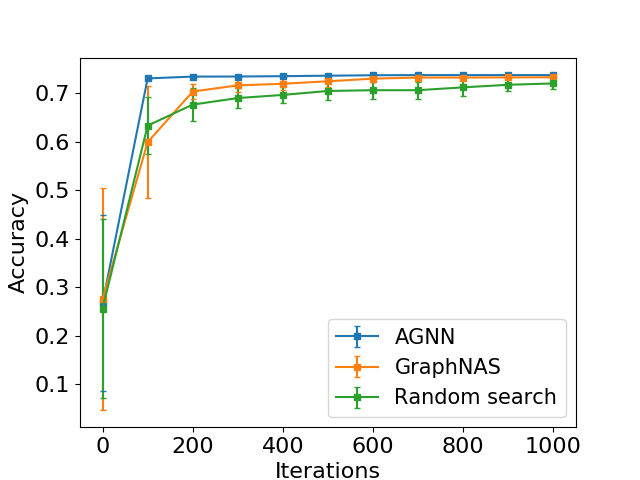}
\end{minipage}}
\subfigure[Pubmed]{
\begin{minipage}{4.27cm}
\centering
\includegraphics[width=4.8cm]{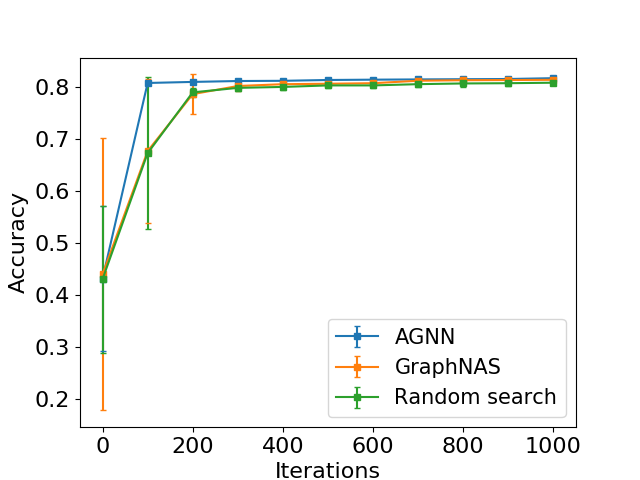}
\end{minipage}}
\vspace{-.2cm}
\caption{Progression of top-$10$ averaged performance of different search methods, i.e., AGNN, GraphNAS, and random search.}
\label{fig:search_effic}
\end{figure*}

\subsubsection{\textbf{Search Efficiency Comparison}}
We compare the progression of top-$10$ averaged performance of our AGNN, GraphNAS and random search, in order to provide positive answer to the research question \textbf{Q2}. All of the search methods are performed without parameter sharing to only study the efficiencies of different controllers. For each search method, totally $1000$ architectures are explored in the same search space. The progression comparisons on the four datasets are shown in Figure \ref{fig:search_effic}. 

As can be seen from Figure \ref{fig:search_effic}, AGNN is more efficient to find the well-performed architectures during the search progress. The top-$10$ architectures discovered by AGNN have better averaged performance on PPI and Citeseer. It is because the best architecture found so far is retained and prepared for slight architecture modification in the next step. Only some actions are resampled to generate the offspring architecture. This will accelerate the search progress toward the better neural architectures among the offsprings. 

\subsubsection{\textbf{Effectiveness Validation of Parameter Sharing}}
\begin{figure*}
\centering
\hspace{-.3cm}
\subfigure[PPI]{
\begin{minipage}{4.27cm}
\centering
\includegraphics[width=4.8cm]{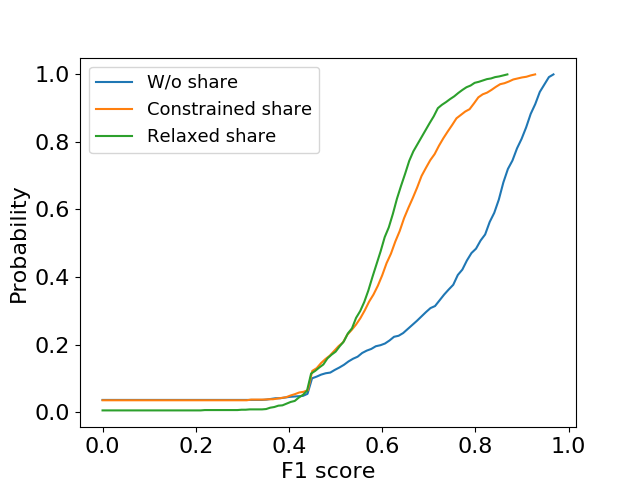}
\end{minipage}}
\subfigure[Cora]{
\begin{minipage}{4.27cm}
\centering
\includegraphics[width=4.8cm]{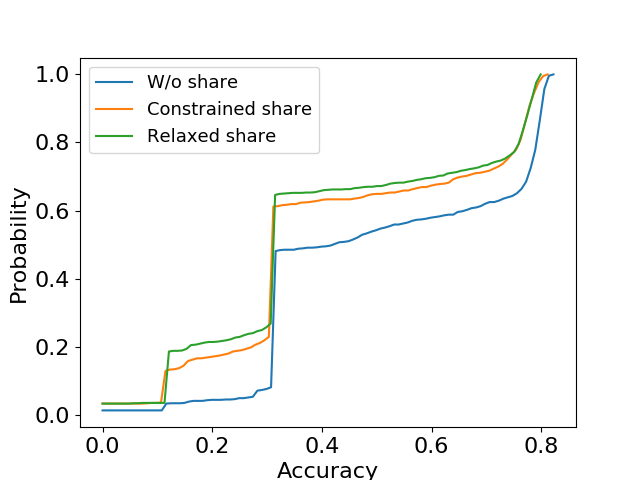}
\end{minipage}}
\subfigure[Citeseer]{
\begin{minipage}{4.27cm}
\centering
\includegraphics[width=4.8cm]{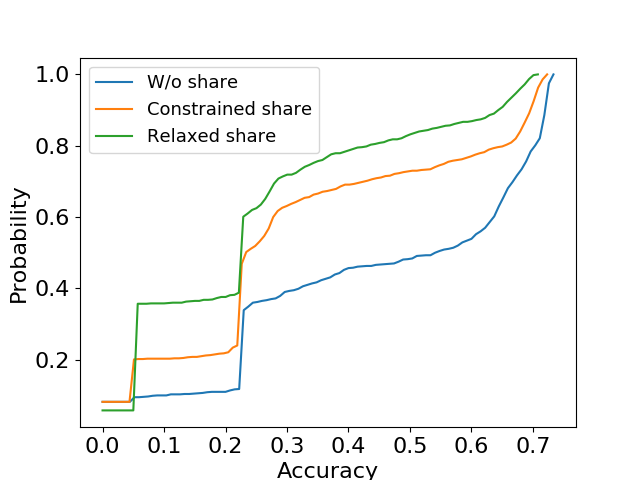}
\end{minipage}}
\subfigure[Pubmed]{
\begin{minipage}{4.27cm}
\centering
\includegraphics[width=4.8cm]{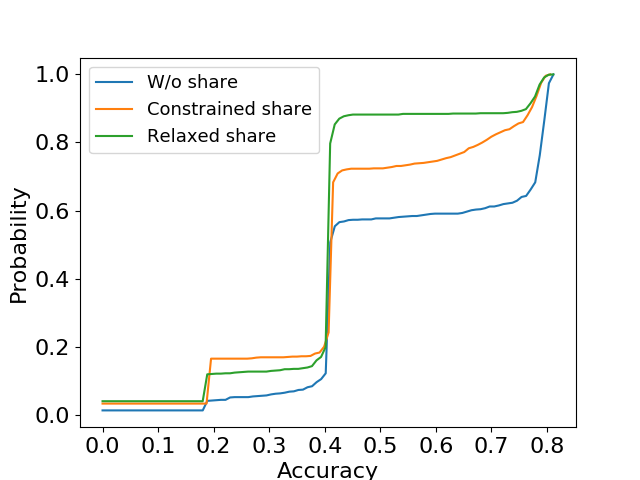}
\end{minipage}}
\vspace{-.35cm}
\caption{The cumulative distribution of validation performance for AGNN under different parameter sharing strategies: the proposed constrained one, the relaxed one in GraphNAS, and training from scratch without parameter sharing.}
\label{fig:paras_share}
\end{figure*}

Herein we study whether or not the shared parameter could be effective in the offspring architecture to help achieve good classification performance, aiming to provide answer for research question \textbf{Q3}. We consider AGNN equipped with different parameter sharing strategies: the proposed constrained one, the relaxed one in GraphNAS, and training from scratch without parameter sharing. Note that the relaxed parameter sharing in GraphNAS is similar to that in the previous NAS framework, at which the offspring architecture shares weight of the same shape directly without any constraint. The cumulative distribution of validation performance is compared for the $1000$ discovered architectures in Figure \ref{fig:paras_share}. 

As can be seen from Figure \ref{fig:paras_share}, most of the neural architectures found by the constrained parameter sharing have better performance than those found by relaxed strategy. That is because the manually-designed constraints limit the parameter sharing only between the homogeneous architectures with similar output statistics. Combined with a few epochs to warm up weight in batch normalization and skip connection, the shared parameter could be effective to the newly sampled architecture. In addition, the offspring architecture is generated with slight architecture modification to the best architecture found so far, which means that they potentially have the similar architecture and output statistics. Hence the well-trained weight could be transferred to the offspring architecture stably. Although the strategy of training from scratch couples the weight to each architecture perfectly, it needs to pay much more computation cost. 
\begin{figure*}
\centering
\hspace{-.3cm}
\subfigure[PPI]{
\begin{minipage}{4.27cm}
\centering
\includegraphics[width=4.8cm]{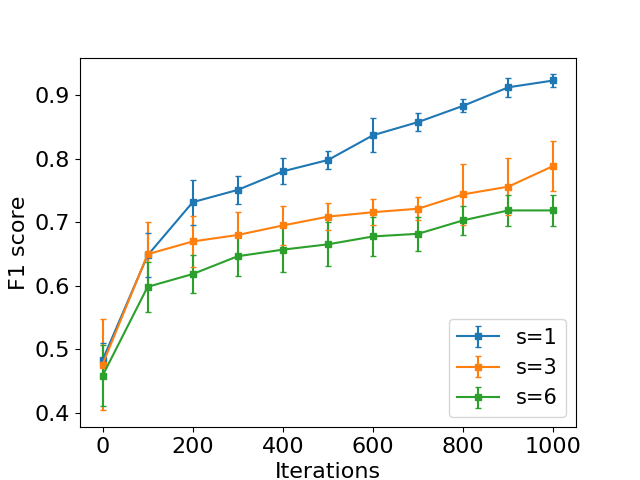}
\end{minipage}}
\subfigure[Cora]{
\begin{minipage}{4.27cm}
\centering
\includegraphics[width=4.8cm]{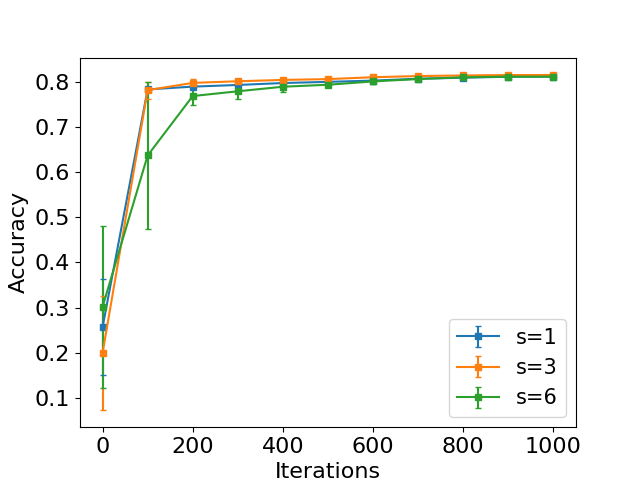}
\end{minipage}}
\subfigure[Citeseer]{
\begin{minipage}{4.27cm}
\centering
\includegraphics[width=4.8cm]{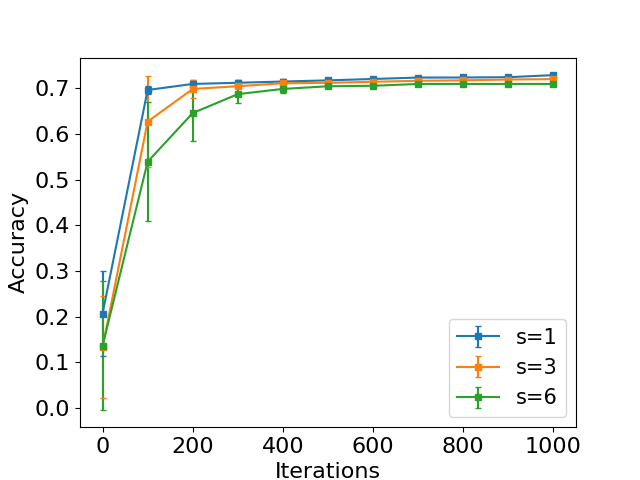}
\end{minipage}}
\subfigure[Pubmed]{
\begin{minipage}{4.27cm}
\centering
\includegraphics[width=4.8cm]{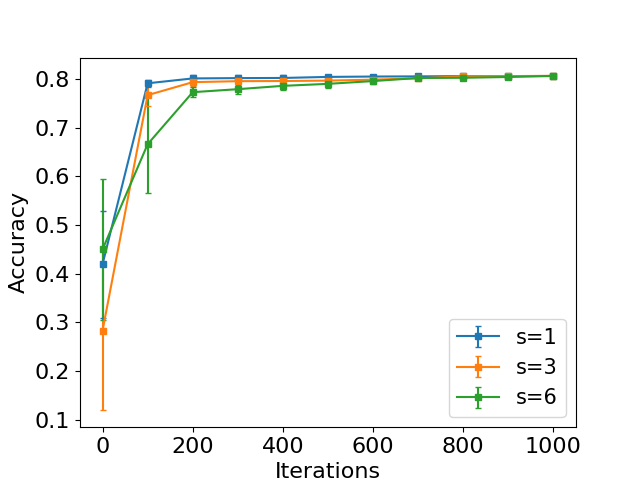}
\end{minipage}}
\vspace{-.35cm}
\caption{The progression of top-$10$ averaged performance of AGNN under different architecture modification: $s=1,3$, and $6$.}
\label{fig:arch_modify}
\end{figure*}
\subsubsection{\textbf{Influence of Architecture Modification}}
We study how does different scales of architecture modification affect the search efficiency, in order to provide answer to research question \textbf{Q4}. Note that the action class in list $\mathcal{C}$ are exploited to modify the retained architecture, and the size of list $\mathcal{C}$ is denoted by $s$. When $s=1$, we perform the architecture modification at the minimum level, at which actions of one specific class will be resampled. When $s=6$, we modify the retained network completely similar to the traditional controller. Considering $s = 1$, $3$, and $6$, we show the progression of top-$10$ architectures under the setting of parameter sharing in Figure \ref{fig:arch_modify}.

As can be seen from Figure \ref{fig:arch_modify}, the architecture search progress tends to be more efficient with the decrease of $s$. The top-$10$ neural architectures found by $s=1$ achieves the best averaged performance on PPI and Citeseer. The efficient progression of smaller $s$ benefits from the following two facts. First, the offspring architecture tends to have similar structure and output statistics with the retained one. It is more possible for the shared weight being effective in the offspring architecture. Second, the independent RNN encoder can exactly learn causality between performance variation and architecture modification of its own class, and tends to sample well-performed architecture at the next step. 

\section{Related Work}
Our work is related to the graph neural networks and neural architecture search.

\textbf{Graph Neural Networks.} A wide variety of GNNs have been proposed to learn the node representation effectively, e.g., recursive neural networks \cite{gori2005new, scarselli2009graph}, graph convolutional networks \cite{bruna2013spectral, defferrard2016convolutional, kipf2016semi, hamilton2017inductive, gao2018large} and graph attention networks \cite{velickovic2017graph,vaswani2017attention}. Most of these approaches are built up based on message-passing based graph convolutions. The underlying graph is viewed as a computation graph, at which node embedding is generated via message passing, information transformation, neighbor aggregation and self update.

\textbf{Neural Architecture Search.} Most of NAS frameworks are built up based on one of the two basic algorithms: RL \cite{zoph2016neural, zoph2018learning, pham2018efficient, cai2017reinforcement, baker2016designing} and EA \cite{liu2017hierarchical, real2017large, miikkulainen2019evolving, xie2017genetic, real2019regularized}. For the former one, a RNN controller is applied to specify the variable-length strings of neural architecture. Then the controller is updated with policy gradient after evaluating the sampled architecture on validation set. For the latter one, a population of architectures are initialized first and evolved with mutation and crossover. The architectures with competitive performance will be retained during the search progress. A new framework combines these two search algorithms to improve the search efficiency \cite{chen2018reinforced}. 
Parameter sharing \cite{pham2018efficient} is proposed to transfer the well-trained weight before to a sampled architecture, to avoid training the offspring architecture from scratch to convergence. 

\section{Conclusion}
In this paper, we present AGNN to find the optimal neural architecture given a node classification task. The search space, RCNAS controller and constrained parameter sharing strategy together are designed specifically suited for the message-passing based GNN. Experiment results show the discovered neural architectures achieve quite competitive performance on both transductive and inductive learning tasks.
The proposed RCNAS controller search the well-performed architectutres more efficiently, and the shared weight could be effective in the offspring network under constraints. For future work, first we will try to apply AGNN to discover architectures for more applications such as graph classification and link prediction. Second, we plan to consider more advanced techniques of graph convolutions in the search space, to facilitate neural architecture search in different applications. 

\balance
\bibliographystyle{unsrt}
\bibliography{reference}

\end{document}